\documentclass[conference]{IEEEtran}
\IEEEoverridecommandlockouts

\usepackage{cite}
\usepackage{amsmath,amssymb,amsfonts}
\usepackage{graphicx}
\usepackage{xcolor}
\usepackage{url}
\usepackage{orcidlink}

\begin{document}

\title{Designing Touch for Trauma-Informed Social Robots: A Design Space for Direct and Indirect Actuation}

\author{
\IEEEauthorblockN{Madeleine Rischer \orcidlink{0009-0004-1471-2452}}
\IEEEauthorblockA{
AI Ethics Research Hub\\
Helmut Schmidt University\\
Hamburg, Germany\\
madeleine.rischer@hsu.hamburg
}
\and
\IEEEauthorblockN{Benedikt Bußmann \orcidlink{0009-0006-2664-3145}}
\IEEEauthorblockA{
AI Ethics Research Hub\\
Helmut Schmidt University\\
Hamburg, Germany\\
benediktbussmann@hsu.hamburg
}
}

\maketitle

\begin{abstract}
Touch is a fundamental communication modality in human-robot interaction and may support grounding, emotional regulation, and stress reduction in therapeutic contexts. However, designing touch-based interactions for individuals with post-traumatic stress disorder (PTSD) requires careful consideration of trauma-informed care (TIC) principles, including safety, transparency, autonomy, and trigger avoidance. This paper investigates how touch actuation in social robots can be designed to align with trauma-informed care. We distinguish between direct touch, involving physical contact between a user and a robot, and indirect touch, mediated through artifacts such as wearables or smart textiles. Building on this distinction, we propose a design space comprising three dimensions: Actuation Modality, Objective of the Actuation, and Intended Effect of the Actuation. We analyze these dimensions through the lens of trauma-informed care and derive key design considerations for touch-based interventions. The resulting approach provides a conceptual foundation for developing trauma-sensitive social robots that support individuals with PTSD through touch-based interactions.
\end{abstract}

\begin{IEEEkeywords}
Touch, Social Robots, Trauma-Informed Design, Tactile Interaction, HRI
\end{IEEEkeywords}

\section{Introduction}
\label{sec:introduction}

Touch represents a fundamental mode of human communication that extends beyond verbal interaction. Within human-robot interaction (HRI), tactile interaction has increasingly attracted attention as a means of creating more embodied, intuitive, and emotionally meaningful forms of engagement between humans and robots \cite{Tsirka.2025}. Social robots can communicate not only through speech, visual cues, and movement, but also through haptic, thermal, and other bodily forms of feedback \cite[cf. p. 972]{Lee.2006} \cite[cf. p. 9]{Ramirez.2022}. Such body-centered interaction modalities are particularly relevant in contexts where emotional regulation and social support are central goals \cite{Coronado.2021} \cite{Bethel.2018} \cite{Laban.2021}.

Touch is also highly relevant in the treatment of individuals with Post-Traumatic Stress Disorder (PTSD) \cite{Kuhfu.2021}. Contemporary PTSD interventions increasingly emphasize physiology-informed and body-centered micro-interventions that support grounding, emotional regulation, stabilization, and stress reduction \cite{Polusny.2015} \cite{Kuhfu.2021}. These approaches may help individuals reconnect with the present moment during periods of heightened arousal or distress and can function as social signals of safety that reduce stress and physiological activation \cite{Eckstein.2020} \cite{Polusny.2015}.

Motivated by the potential of social robots to provide supportive therapeutic micro-interventions for individuals with PTSD in everyday situations \cite[cf. p. 4]{Laban.2021}, this work investigates touch as a communication and actuation modality within trauma-sensitive human-robot interaction.

However, designing touch-based interaction for PTSD support introduces unique challenges. Trauma-informed care (TIC) emphasizes core principles such as safety, trustworthiness, transparency, empowerment, choice, and sensitivity toward individual trauma triggers \cite[cf. p. 6]{Mazuz.2025}. Touch that is perceived as unexpected, invasive, or reminiscent of traumatic experiences may cause distress or contribute to re-traumatization \cite{Mazuz.2025}. Consequently, touch-based interventions must be designed to respect autonomy, consent, individual boundaries, and user control over bodily interaction \cite{Mazuz.2025}.

These considerations lead to the following research question:

\textbf{RQ}: How should touch actuation for social robots be designed to align with trauma-informed care principles?

To address this question, the paper distinguishes between direct and indirect touch as two forms of robot-mediated tactile interaction and develops a design space consisting of three dimensions for characterizing touch actuation. This design space is then analyzed through the lens of trauma-informed care to derive key considerations for designing trauma-sensitive touch interactions in future social robots for PTSD support.

\section{Direct and Indirect Touch}
\label{sec:direct_indirect_touch}

Having established the need for trauma-sensitive touch design in social robots, it is first necessary to clarify the different ways touch can occur in human-robot interaction. Existing HRI research shows that touch-based interaction is not limited to direct physical contact between a robot and a user. Instead, tactile interaction may also be mediated through technological artifacts [citation
omitted because of ongoing blind review] \cite[cf. p. 7]{Ruckdashel.2022} \cite{Papadopoulou.2019}. We therefore distinguish between \emph{direct touch} and \emph{indirect touch}.

\textbf{Direct touch} refers to tactile interaction in which physical contact occurs directly between the human and the robot, without an intermediary device. It includes both human-to-robot and robot-to-human contact. Humans may touch robots through actions such as stroking, petting, or holding them, as commonly observed in companion and therapeutic robotics; for example, users of Therabot™ engage with the robot through tactile interaction with its fur-like surface, while the Ommie robot supports guided breathing exercises through direct contact with its inflatable body \cite{Bethel.2018} \cite{Matheus.2025}.

Conversely, robots may touch humans by initiating physical contact through behaviors such as poking, leaning, or holding. Hugging robots represent a prominent example of this interaction form \cite{Block.2023}.

By contrast, \textbf{indirect touch} refers to tactile interaction mediated through an additional technological artifact.

In this case, a robot may influence a user through intermediary devices such as wearables, smart textiles, or other body-worn systems that generate tactile, thermal, or pressure-based stimuli \cite[cf. pp. 7, 13]{Ruckdashel.2022}. Rather than touching the user directly, the robot controls the actuation of these devices. Our previously proposed prototype [citation omitted because of ongoing blind review], for example, combines a social robot with smart textile actuators that provide haptic feedback on the user's body.

Alternatively, users may provide tactile input through intermediary devices such as buttons, wearable controllers, pressure-sensitive interfaces, or capacitive sensors integrated into garments \cite[cf. p. 5]{Ruckdashel.2022}. In our prototype, users can accept or decline robot-initiated interventions through capacitive textile sensors embedded in a smart garment [citation omitted because of ongoing blind review].

Indirect touch extends tactile interaction beyond the robot's physical body and enables the integration of wearable and textile-based technologies into HRI. This distinction is particularly relevant in trauma-informed contexts. While direct touch may offer natural and emotionally meaningful interactions, indirect touch can provide greater flexibility, controllability, personalization, and physical distance. These characteristics may be especially valuable when designing for individuals with PTSD, where bodily boundaries, consent, and trigger avoidance require careful consideration \cite{Mazuz.2025}.

\section{Design Space for Touch Actuation}
\label{sec:design_space}

While the distinction between direct and indirect touch describes how tactile interaction is physically established between a human and a robot, it does not capture the variety of ways in which touch can be implemented and experienced. Touch-based interventions can differ substantially regarding the sensory modality employed, the bodily target of the intervention, and the intended outcome. To systematically characterize these differences, we propose a design space consisting of three dimensions: \emph{Actuation Modality}, \emph{Objective of the Actuation}, and \emph{Intended Effect of the Actuation}. Together, these dimensions provide a structured approach for analyzing and designing touch-based interactions in social robotics.

\subsection{Dimension 1: Actuation Modality}
\label{sec:Dim1}

The first dimension describes the physical mechanism through which tactile stimulation is generated and encompasses vibration, thermal and airflow actuation, as well as stretch and pressure \cite[cf. pp. 7, 13]{Ruckdashel.2022}.

Vibrotactile feedback is among the most common forms of touch actuation in wearable technologies and human-computer interaction \cite[cf. p. 105]{Hong.2023}. \textbf{Vibration} can be delivered through actuators embedded in robots, wearables, or smart textiles and can vary in intensity, frequency, duration, and rhythm \cite[cf. p. 105]{Hong.2023}. Typical application areas include gaming, robot teleoperation and Virtual Reality \cite[cf. p. 108]{Hong.2023}. For example, our previously proposed prototype utilized vibrating motor discs embedded within a smart textile to guide breathing exercises through rhythmic haptic cues [citation omitted because of ongoing blind review].

\textbf{Thermal and airflow actuation} influence the user through changes in temperature reception \cite{Raza.2025}. Thermal actuation may involve heating or cooling elements integrated into robots, wearable devices, or textiles \cite{Raza.2025}. Airflow-based actuation can be realized through fans \cite{Lee.2016}.

\textbf{Stretch and pressure}-based actuation physically deform the skin, clothing, or surrounding environment to create tactile sensations. Examples include compression garments, capacitive touch sensors, or robotic hugs \cite{Papadopoulou.2019} \cite[cf. p. 5]{Ruckdashel.2022} \cite{Block.2023}. Pressure-based interventions may simulate holding, squeezing, guiding, or comforting touch and therefore represent one of the most direct forms of tactile stimulation available in human-robot interaction. A prominent example is HuggieBot 3.0, illustrating how pressure-based actuation can be employed to emulate supportive interpersonal touch while maintaining control over force and user comfort \cite[cf. p. 40]{Block.2023}.

\subsection{Dimension 2: Objective of the Actuation}
\label{sec:Dim2}

The second dimension describes the primary target of the intervention within the human body. We distinguish between interventions directed at the \emph{body as a whole}, the \emph{external body}, and the \emph{internal body}.

\textbf{Whole-body} interventions aim to influence the overall bodily state without targeting a specific anatomical structure. Examples include full-body cooling systems or vibration motors across the whole body \cite{Raza.2025} \cite{Park.2025}. The primary objective is often the modulation of global arousal, relaxation, or bodily awareness \cite{Reynolds.2015} \cite{Eckstein.2020}.

\textbf{Externus}-oriented interventions primarily target the external body, particularly the skin, and body surface. Examples include thermal stimulation applied to the skin or robotic petting \cite{Papadopoulou.2019} \cite{Bethel.2018}. Such interventions directly stimulate bodily receptors and are typically perceived as external tactile events \cite{Huang.2023}.

\textbf{Internus}-oriented interventions seek to influence internal physiological processes or bodily functions. Although the actuation itself is applied externally, it serves as a mechanism for affecting internal bodily states rather than merely producing external sensations. Examples include breathing control through rhythmic vibration or pressure [citation omitted because of ongoing blind review] \cite{Matheus.2025}.

\subsection{Dimension 3: Intended Effect of the Actuation}
\label{sec:Dim3}

The third dimension describes the primary purpose of the touch-based intervention.

\textbf{Informational touch} communicates information to the user. The tactile stimulus functions as a communication channel analogous to visual or auditory feedback \cite[cf. p. 972]{Lee.2006}. Examples may include notifications, warnings, directional cues, reminders, or indications of robot states. 

\textbf{Functional touch} directly supports the execution of a task or behavior. Here, the tactile feedback is intended to guide, structure, or facilitate an action. For example, rhythmic vibrations or inflation–deflation patterns can be used to support breathing exercises [citation omitted because of ongoing blind review] \cite{Matheus.2025}.

\textbf{Affective touch} aims to influence emotional experiences, feelings, or social perception. Examples include comforting warmth, soothing tactile stimulation like petting, or robotic hugs \cite{Papadopoulou.2019} \cite{Bethel.2018} \cite{Block.2023}. Affective touch is particularly relevant in therapeutic and social robotics contexts because it may contribute to emotional regulation, stress reduction, and perceived social support \cite{Eckstein.2020} \cite{Bethel.2018} \cite{Coronado.2021}.

These three dimensions are not mutually exclusive. A single touch-based intervention may simultaneously involve multiple actuation modalities, target both external and internal bodily processes, and pursue informational, functional, and affective goals. Nevertheless, the proposed design space provides a structured vocabulary for describing and comparing touch-based interaction concepts. In the following section, we examine how the different regions of this design space align with TIC principles and discuss the opportunities and risks associated with each design choice.

\section{Trauma-Informed Considerations}
\label{sec:trauma_informed}

Individuals with PTSD may exhibit heightened sensitivity toward bodily sensations, physical proximity, and environmental stimuli, making the design of touch-based interventions particularly challenging \cite[cf. p. 769 f.]{Pitman.2012}. Consequently, each design dimension should be considered through the lens of TIC principles.

The selected actuation modalities (section~\ref{sec:Dim1}) directly determine how touch is physically experienced and therefore influence perceived safety, trust, and emotional comfort.

A primary consideration is the TIC principle of \textbf{Safe Relationship and Safe Environment} \cite[cf. pp. 6, 8]{Mazuz.2025}. Actuation modalities should neither expose users to physical harm nor create a perception of danger \cite{Mazuz.2025}. Thermal actuators, for example, must remain within safe temperature ranges, while pressure- and vibration-based systems should avoid excessive force or overwhelming stimulation. Beyond objective safety, designers must also consider perceived safety, as actuators that appear capable of causing harm may undermine trust even when no actual risk exists \cite[cf. p. 1]{Rubagotti.2022}.

Closely related is \textbf{Trigger Avoidance} \cite[cf. p. 9]{Mazuz.2025}. Certain tactile sensations may unintentionally evoke traumatic memories or associations \cite{Mazuz.2025} \cite{Franke.2021}. Sudden pressure, unexpected vibrations, or rapidly changing thermal stimuli may be interpreted as threatening depending on an individual's trauma history. Designers should therefore favor gradual, predictable, and user-adjustable forms of actuation.

The TIC principle of \textbf{Trustworthiness and Transparency} requires tactile systems to behave consistently and predictably \cite[cf. p. 6 f.]{Mazuz.2025}. Users should be able to anticipate how an actuator will respond and what sensation it will generate. Unexpected or unexplained behavior may reduce trust and perceived reliability.

Finally, \textbf{Empowerment, Choice, and Control} are particularly important in touch-based interaction \cite[cf. pp. 6 f., 9]{Mazuz.2025}. Users should understand how an actuator functions, when it is activated, and how it can be interrupted or disabled \cite{Mazuz.2025}. Explicit consent requests, adjustable intensity levels, and accessible stop functions can help preserve autonomy and reinforce a sense of control.

The location(s) and bodily target(s) of touch-based interventions (section~\ref{sec:Dim2}) are similarly important from a trauma-informed perspective. While some body locations may be experienced as neutral or supportive, others may be associated with vulnerability, discomfort, or traumatic experiences \cite{Mazuz.2025}.

Moreover, \textbf{Trustworthiness and Transparency} require users to clearly understand where touch will occur and which body regions may be affected. For example, users should know whether a vibration motor embedded in a garment stimulates the chest, shoulder, arm, or another region.

At the same time, \textbf{Trigger Avoidance} remains central. Trauma survivors may react differently to touch at specific body locations. Sensitive areas such as the chest, neck, back, or abdomen may evoke unwanted emotional responses in some individuals. Actuator placement should therefore be carefully considered and, whenever possible, individualized to accommodate personal preferences and sensitivities.

The TIC principle of \textbf{Survivor Partnerships} further suggests that intervention objectives should be selected according to therapeutic goals and individual needs rather than technical considerations alone \cite[cf. p. 6]{Mazuz.2025}. Different micro-interventions may benefit from targeting different bodily regions, making collaboration with therapists and users essential.

From a trauma-informed perspective, the intended effect(s) of touch-based interaction (section~\ref{sec:Dim3}) should be guided by principles that emphasize user-centered goals, meaningful support, and collaborative interaction.

Firstly, the principle of \textbf{Survivor Partnerships} is particularly relevant because desired effects should align with users' micro-therapeutic goals and preferences. Touch should serve meaningful outcomes such as grounding, emotional regulation, stress reduction, or relaxation rather than being implemented solely because it is technically feasible.

Secondly, \textbf{Providing Meaningful Companionship} highlights the importance of ensuring that touch contributes to the robot's supportive role in everyday life \cite[cf. pp. 6 ff.]{Mazuz.2025}. Whether informational, functional, or affective, tactile interventions should support well-being, contribute to therapeutic goals, and reinforce the robot's role as a trustworthy source of support \cite{Mazuz.2025}.

Lastly, \textbf{Collaboration and Mutuality} emphasize that touch should be integrated into a mutually understandable interaction \cite[cf. pp. 6 ff.]{Mazuz.2025}. Informational touch should clearly communicate information, functional touch should support task execution or micro-therapeutic interventions, and affective touch should align with the social and emotional context. Misalignment between communicative intent and user interpretation may lead to confusion, reduced trust, or ineffective interaction \cite[cf. p. 440]{Schramm.2020}.

\begin{figure}[ht]
    \centering
    \includegraphics[width=\linewidth]{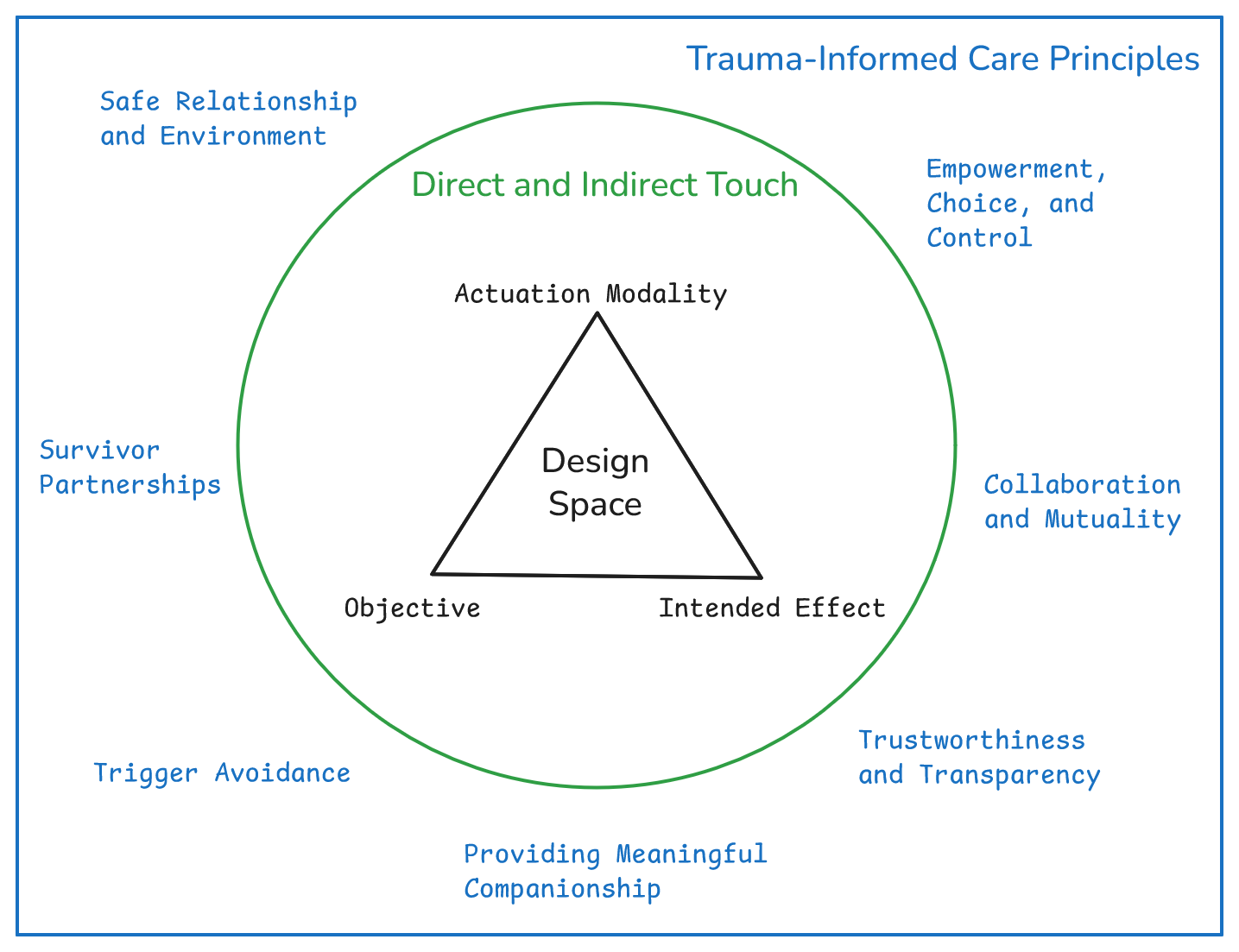}
    \caption{Proposed approach for trauma-informed social touch actuation. Design considerations are characterized by actuation modality, objective, and intended effect, situated within the broader spectrum of direct and indirect touch modalities and framed by TIC principles that support safe, transparent, collaborative, and empowering touch interactions.}
    \label{fig:1}
\end{figure}

Taken together, these considerations demonstrate that trauma-informed touch design extends beyond the selection of a particular actuator or interaction technique. Rather, trauma-informed touch emerges from the careful alignment of actuation modalities, intervention objectives, and intended effects with the principles of safety, transparency, autonomy, collaboration, and trigger sensitivity. The proposed design space therefore not only supports the systematic description of touch-based interactions but also provides an approach for evaluating their suitability within trauma-informed social robotics. Fig.~\ref{fig:1} synthesizes the overall approach by linking the distinction between direct and indirect touch, the proposed design space, and the TIC principles that guide the design and evaluation of touch-based interventions.

\section{Conclusion and Future Work}
\label{sec:conclusion}

This short paper examined how touch actuation in social robots can be designed in accordance with TIC principles.

We distinguished between direct touch, involving physical contact with the robot, and indirect touch, mediated through artifacts such as wearables or smart textiles. This distinction is important because direct and indirect touch offer different levels of autonomy, consent, and bodily engagement, which are central considerations in trauma-informed care. Building on this, we proposed a design space consisting of three dimensions, Actuation Modality, Objective of the Actuation, and Intended Effect of the Actuation, which provides a structured approach for describing and comparing touch-based interactions in social robotics.

Our analysis suggests that touch actuation can support trauma-informed care when it is designed with attention to safety, transparency, trustworthiness, user autonomy, trigger avoidance, collaboration, and survivor partnerships. Touch should therefore be considered not only from a technical perspective but also in terms of its psychological and therapeutic implications, particularly its ability to enable meaningful and user-controlled therapeutic micro-interventions.

More broadly, this work provides an initial conceptual foundation for trauma-informed touch design in social robotics, especially for vulnerable populations such as individuals with PTSD. However, the proposed design space remains exploratory and has not yet been empirically evaluated with PTSD patients, therapists, or other stakeholders. As a result, no conclusions can currently be drawn regarding the effectiveness, acceptance, or clinical suitability of specific touch-based interventions.

Future research should therefore examine and refine the design space through participatory and empirical studies involving PTSD patients, therapists, and other stakeholders. Particular attention should be given to potentially triggering stimuli, preferred body locations, and acceptable levels of intensity and autonomy. Further work should also investigate how PTSD-related symptoms and therapeutic goals can be mapped to suitable touch modalities, bodily targets, and intended effects, as well as compare direct and indirect touch approaches regarding user experience, perceived safety, trust, and therapeutic outcomes.

Finally, this work is part of a broader research agenda on trauma-informed social robots for PTSD support. The overall goal is to develop interaction strategies that are technically feasible, ethically appropriate, and aligned with the needs of trauma survivors.

\section*{AI Usage Acknowledgments}

 Any use of generative AI in this paper adheres to ethical guidelines for use and acknowledgment of generative AI in academic research, as outlined in \cite{PorsdamMann.2024}. The authors have made a substantial contribution to the work, which has been thoroughly vetted for accuracy, and assume responsibility for the integrity of the contribution.
\textsf{ChatGPT-5.5}, and \textsf{DeepL Write} were used for rephrasing and stylistic refinement of individual paragraphs. \textsf{DeepL Translator} was used for translating individual phrases.


\bibliographystyle{IEEEtran}
\bibliography{references}

\end{document}